\documentclass[english]{article}
\pdfoutput=1
\usepackage{amssymb}
\usepackage{amsmath}
\usepackage{wrapfig}
\usepackage{graphicx}
\usepackage{subfigure}
\usepackage{verbatim}
\usepackage[multiple]{footmisc}
\usepackage[bf]{caption2}

\graphicspath{{images/}}
\newcommand{\eg}{e.g.\ }

\begin{document}

\title{Comparison and Combination of State-of-the-art \mbox{\rule{-1ex}{0pt}Techniques for 
Handwritten Character Recognition:} Topping the MNIST Benchmark}
\author{Daniel Keysers\\daniel.keysers@dfki.de\\
Image Understanding and Pattern Recognition (IUPR) Group\\
German Research Center for Artificial Intelligence (DFKI)}
\date{May 2006}
\maketitle

\pagestyle{plain}
\setcounter{page}{1}

\begin{abstract}
  Although the recognition of isolated handwritten digits has been a research
  topic for many years, it continues to be of interest for the research
  community and for commercial applications.  We show that despite the
  maturity of the field, different approaches still deliver results that vary
  enough to allow improvements by using their combination.  We do so by
  choosing four well-motivated state-of-the-art recognition systems for which
  results on the standard MNIST benchmark are available. When comparing the
  errors made, we observe that the errors made differ between all four
  systems, suggesting the use of classifier combination. We then determine the
  error rate of a hypothetical system that combines the output of the four
  systems. The result obtained in this manner is an error rate of 0.35\% on
  the MNIST data, the best result published so far. We furthermore discuss the
  statistical significance of the combined result and of the results of the
  individual classifiers.
\end{abstract}

\section{Introduction}

The recognition of handwritten digits is a topic of practical importance
because of applications like automated form reading and handwritten zip-code
processing. It is also a subject that has continued to produce much research 
effort over the last decades for several reasons: 
\begin{itemize}
\addtolength{\itemsep}{-1.2ex}
\item The problem is prototypical for image processing and pattern recognition, with
a small number of classes. 
\item Standard benchmark data sets exist that make it easy to obtain valid results
quickly. 
\item Many publications and techniques are available that can be cited and
  built on, respectively.
\item The practical applications motivate the research performed.
\item Improvements in classification accuracy over existing techniques
  continue to be obtained using new approaches.
\end{itemize}

This paper has the objective to analyze four of the state-of-the-art methods
for the recognition of handwritten
digits~\cite{shapecontext_pami,sch02,icpr04_nlmatch,simardICDAR03} by
comparing the errors made on the standard MNIST benchmark data. 
(A part of this work has been described in~\cite{diss}.)
We perform a
statistically analysis of the errors using a bootstrapping
technique~\cite{bisani_poi} that not only uses the error count but also takes
into account which errors were made. Using this technique we can determine
more accurate estimates of the statistical significance of improvements.

When analyzing the errors made we observe that --- although the error rates
obtained are all very similar --- there are substantial differences in {\em
  which} patterns are classified erroneously.  This can be interpreted as an
indicator for using classifier combination. An experiment shows that indeed a
combination of the classifiers performs better than the single best
classifier. The statistical analysis shows that the probability that this
results constitutes a real improvement and is not based on chance alone is
94\%.

\section{Related work} 

This paper is of course only possible because the results
of the four chosen base
methods~\cite{shapecontext_pami,sch02,icpr04_nlmatch,simardICDAR03} were
available\footnote{We would like to thank Patrice Simard for providing the
  recognition results to us and the authors of~\cite{shapecontext_pami,sch02}
  for listing the errors in the respective papers.}.  These approaches are
presented in more detail in Section~\ref{sec:mnist-sota}.  We are aware that
there exist other methods that also achieve very good classification error
rates on the data used, e.g.~\cite{liu_benchmark}. 
However, we feel that the
four methods chosen comprise a set of well-motivated and self-contained
approaches. 
Furthermore, they represent the different classification methods
most commonly used (in the research literature), that is, the nearest neighbor
classifier, neural networks, and the support vector machine.  All four methods
use the appearance-based paradigm in the broad sense and can thus be
considered as being sufficiently general as to be applied to other object
recognition tasks. 

There is a large amount of work available on the topic of classifier
combination as well (an introduction can be found e.g.~in~\cite{Kittler98})
and much work exists on applying classifier combination to handwriting
recognition
(e.g.~\cite{batthacharyya-iwfhr04,das06_kumar,mcs2001,bunke-iwfhr04}).
Note that we do not propose new algorithms for classification of handwritten
digits or for the combination of classifiers. Instead, our contribution is to
present a statistical analysis that compares different classifiers and to show
that their combination improves the performance even though the individual
classifiers all reach state-of-the-art error rates by themselves.

\section{The MNIST task}\label{sec:mnist}

\begin{figure}[tb]
\begin{center}
\includegraphics[width=\columnwidth]{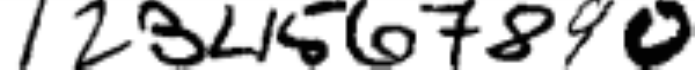}
\includegraphics[width=\columnwidth]{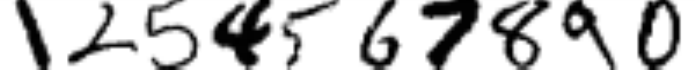}
\includegraphics[width=\columnwidth]{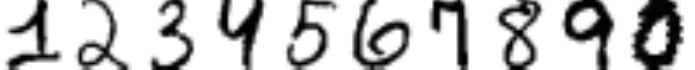}\\
\caption[Example images from the MNIST data set]%
{Example images from the MNIST data set.\label{fig:nist_ex}}
\end{center}
\end{figure}

The modified NIST handwritten digit database (MNIST, \cite{lecun98}) contains
60,000 images in the training set and 10,000 patterns in the test set, each of
size 28$\times$28 pixels with 256 graylevels.  The data set is available
online\footnote{\tt http://www.research.att.com/$\sim$yann/ocr/mnist/} and
some examples from the MNIST corpus are shown in Figure~\ref{fig:nist_ex}.

The preprocessing of the images is described as follows in \cite{lecun98}:
``The original black and white (bilevel) images were size normalized to fit in
a 20$\times$20 pixel box while preserving their aspect ratio. The resulting
images contain gray levels as result of the antialiasing (image interpolation)
technique used by the normalization algorithm. [...]  the images were centered
in a 28$\times$28 image by computing the center of mass of the pixels and
translating the image so as to position this point at the center of the
28$\times$28 field.'' Note that some authors use a `deslanted' version of the
database.

The task is generally not considered to be `difficult' (in the sense that
absolute error rates are high) recognition task for two reasons. First, the
human error rate is estimated to be only about 0.2\%, although it has not been
determined for the whole test set \cite{sim93+}.  Second, the large training
set allows machine learning algorithms to generalize well. With respect to the
connection between training set size and classification performance for OCR
tasks it is argued \cite{Smith94} that increasing the training set size by a
factor of ten cuts the error rate approximately to half the original figure.

Table~\ref{tab:mnist-er} gives a comprehensive overview of the error rates
reported for the MNIST data.  One disadvantage of the MNIST corpus is that
there exists no development test set, which leads to effects known as
`training on the testing data'. This is not necessarily true for each of the
research groups performing experiments, but it cannot always be ruled out.
Note that in some publications (e.g.~\cite{simardICDAR03}) the authors
explicitly state that all parameters of the system were chosen by using a
subset of the training set for validation, which then rules out the
overadaptation to the test set.
However, the tendency exists to evaluate one method with
different parameters or different methods several times on the same data until
the best performance seems to have been reached. This procedure leads to an
overly optimistic estimation of the error rate of the classifier and the
number of tuned parameters should be considered when judging such error rates.
Ideally, a development test set would be used to determine the best parameters
for the classifiers and the results would be obtained from one run on the test
set itself.  Nevertheless a comparison of `best performing' algorithms may
lead to valid conclusions, especially if these perform well on several
different tasks.
\def\pz{\phantom{0}}
\begin{table}[b]
 \caption[MNIST error rates]{Error rates for the MNIST task in \%.
  The systems marked with $^*$ are those we use for analysis and combination.
   \label{tab:mnist-er}}
\centering
\begin{tabular}{@{\vline\hspace{0.7ex}}r@{\hspace{0.7ex}}l@{\hspace{0.7ex}\vline\hspace{0.7ex}}l@{\hspace{0.7ex}\vline\hspace{0.7ex}}r@{\hspace{0.7ex}\vline}}
\hline
\multicolumn{2}{@{\vline\hspace{0.7ex}}l}{reference} & method & ER\pz \\
\hline
\cite{sim93+}             & AT\&T   & human performance & 0.2\pz \\
                          &  ---   & Euclidean nearest neighbor & 3.5\pz \\ 
\hline                     
\cite{maree04}            & U Liège   & decision trees + sub-windows & 2.63 \\
\cite{lecun98}            & AT\&T   & deslant, Euclidean 3-NN  & 2.4\pz \\
\cite{icpr04_uchida}      & Kyushu U & elastic matching & 2.10 \\
\cite{icpr00_td}          & RWTH      & one-sided tangent distance & 1.9\pz \\
\cite{bot94+}             & AT\&T   & neural net LeNet1  & 1.7\pz \\
\cite{mayraz}             & UC London &   products of experts & 1.7\pz \\
\cite{milgram_mnist_05}   & U Qu\'ebec & hyperplanes + support vector m. & 1.5\pz \\
\cite{sch97}              & TU Berlin   & support vector machine & 1.4\pz \\
\cite{bot94+}             & AT\&T      & neural net LeNet4  & 1.1\pz \\
\cite{sim93+}             & AT\&T    & tangent distance  & 1.1\pz \\
\cite{icpr00_td}          & RWTH    & two-sided tangent d., virt. data & 1.0\pz \\
\cite{dong02}             & CENPARMI & local learning & 0.99 \\
\cite{sch98new+}          & MPI, AT\&T   & virtual SVM  & 0.8\pz \\
\cite{lecun98}            & AT\&T   & distortions, neural net LeNet5   & 0.82 \\
\cite{lecun98}            & AT\&T    & distortions, boosted LeNet4          & 0.7\pz \\
\cite{teow00}             & U Singapore  & bio-inspired features + SVM    & 0.72 \\
\cite{sch02}              & Caltech,MPI   & virtual SVM (jitter)         & 0.68 \\
\cite{shapecontext_pami}  & UC Berkeley & shape context matching   & $^*$0.63 \\
\cite{dong04}             & CENPARMI    & support vector machine  & 0.60 \\
\cite{teow02}             & U Singapore  & deslant, biology-inspired features & 0.59 \\
\cite{athi05}             & Boston U & cascaded shape context & 0.58 \\ 
\cite{sch02}              & Caltech,MPI   & deslant, virtual SVM (jitter,shift) & $^*$0.56 \\ 
\cite{athi05}             & Boston U & shape context matching & 0.54 \\ 
\cite{icpr04_nlmatch}     & RWTH  & deformation model (IDM) &  $^*$0.54 \\
\cite{liu_benchmark}      & Hitachi & preprocessing,  support vector m. & 0.42\\
\cite{simardICDAR03}      & Microsoft  & neural net + virtual data & $^*$0.42 \\
                          & this work &  hyp. comb. of 4 systems ($^*$)& 0.35 \\
\hline
\end{tabular}
\end{table}
Note that Dong gives lower error rates than in \cite{dong04} of 0.38 to 0.44
percent on his web page (accessed February 2005), but it remains somewhat
unclear how these error rates were obtained and if possibly these low error
rates are due to the effect of `training on the testing data'.
Also, \cite{teow02} try a variety of SVMs and networks which yield error rates
ranging from 0.59 percent to 0.81 percent. 
The IDM \cite{icpr04_nlmatch} as described in the Section~\ref{sec:idm} was
not optimized for the MNIST task. Instead, all parameter settings were
determined using the smaller USPS data set and then the complete setup was
evaluated once on the MNIST data.

\begin{figure}[p]
\newlength{\mndlength}
\setlength{\mndlength}{7mm}
\newcommand{\examplewithclass}[2]{\scriptsize%
#2\includegraphics[width=\mndlength]{images/MNIST-difficult#1}}
\centerline{
\begin{tabular}{*{8}{c@{\;}}}
\examplewithclass{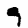}{9}&
\examplewithclass{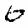}{6}&
\examplewithclass{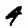}{4}&
\examplewithclass{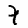}{7}&
\examplewithclass{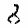}{8}&
\examplewithclass{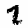}{2}&
\examplewithclass{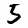}{5}&
\examplewithclass{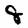}{8}\\
\examplewithclass{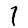}{1}&
\examplewithclass{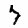}{7}&
\fbox{\examplewithclass{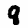}{8}}&
\examplewithclass{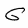}{6}&
\examplewithclass{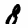}{8}&
\examplewithclass{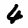}{4}&
\examplewithclass{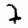}{7}&
\examplewithclass{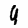}{9}\\
\examplewithclass{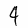}{4}&
\examplewithclass{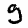}{9}&
\examplewithclass{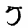}{5}&
\examplewithclass{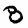}{8}&
\examplewithclass{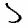}{5}&
\examplewithclass{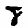}{8}&
\examplewithclass{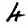}{4}&
\examplewithclass{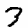}{3}\\
\examplewithclass{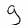}{9}&
\examplewithclass{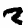}{2}&
\examplewithclass{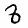}{8}&
\fbox{\examplewithclass{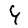}{9}}&
\examplewithclass{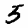}{5}&
\examplewithclass{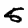}{5}&
\examplewithclass{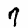}{7}&
\examplewithclass{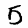}{5}\\
\examplewithclass{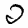}{2}&
\examplewithclass{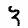}{3}&
\fbox{\examplewithclass{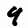}{4}}&
\examplewithclass{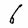}{6}&
\fbox{\examplewithclass{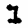}{1}}&
\examplewithclass{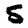}{5}&
\examplewithclass{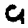}{9}&
\examplewithclass{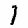}{1}\\
\examplewithclass{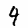}{4}&
\examplewithclass{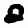}{2}&
\examplewithclass{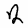}{2}&
\examplewithclass{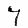}{7}&
\examplewithclass{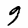}{9}&
\examplewithclass{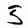}{5}&
\examplewithclass{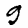}{9}&
\examplewithclass{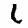}{6}\\
\examplewithclass{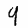}{4}&
\examplewithclass{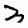}{3}&
\examplewithclass{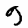}{9}&
\examplewithclass{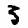}{3}&
\examplewithclass{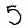}{5}&
\examplewithclass{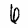}{6}&
\examplewithclass{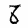}{8}&
\examplewithclass{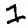}{1}\\
\examplewithclass{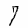}{7}&
\examplewithclass{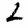}{2}&
\examplewithclass{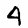}{4}&
\fbox{\examplewithclass{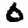}{6}}&
\examplewithclass{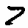}{7}&
\examplewithclass{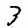}{3}&
\examplewithclass{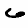}{6}&
\examplewithclass{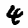}{4}\\
\examplewithclass{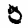}{5}&
\examplewithclass{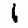}{1}&
\examplewithclass{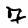}{7}&
\examplewithclass{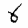}{6}&
\examplewithclass{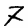}{7}&
\examplewithclass{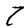}{7}&
\examplewithclass{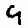}{9}&
\examplewithclass{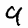}{9}\\
\examplewithclass{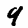}{9}&
\examplewithclass{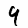}{9}&
\examplewithclass{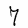}{7}&
\examplewithclass{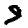}{9}&
\examplewithclass{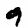}{9}&
\examplewithclass{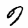}{9}&
\examplewithclass{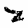}{2}&
\examplewithclass{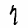}{1}\\
\examplewithclass{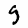}{9}&
\examplewithclass{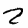}{2}&
\examplewithclass{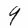}{9}&
\examplewithclass{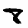}{8}&
\examplewithclass{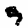}{9}&
\examplewithclass{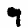}{9}&
\examplewithclass{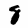}{8}&
\examplewithclass{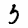}{3}\\
\examplewithclass{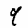}{9}&
\examplewithclass{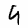}{9}&
\examplewithclass{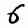}{6}&
\examplewithclass{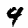}{4}&
\examplewithclass{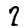}{7}&
\examplewithclass{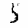}{5}&
\fbox{\examplewithclass{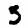}{5}}&
\examplewithclass{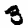}{3}\\
\examplewithclass{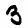}{3}&
\examplewithclass{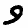}{9}&
\examplewithclass{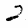}{2}&
\examplewithclass{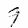}{9}&
\examplewithclass{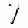}{7}&
\examplewithclass{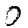}{0}&
\examplewithclass{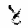}{8}&
\examplewithclass{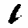}{1}\\
\examplewithclass{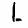}{1}&
\examplewithclass{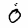}{0}&
\examplewithclass{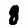}{8}&
\examplewithclass{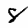}{8}&
\examplewithclass{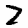}{7}&
\examplewithclass{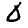}{0}&
\examplewithclass{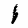}{1}&
\examplewithclass{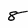}{8}\\
\examplewithclass{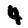}{4}&
\examplewithclass{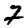}{7}&
\examplewithclass{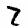}{7}&
\examplewithclass{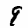}{9}&
\examplewithclass{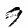}{9}&
\examplewithclass{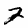}{2}&
\examplewithclass{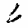}{6}&
\examplewithclass{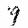}{9}\\
\examplewithclass{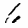}{6}&
\fbox{\examplewithclass{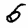}{5}}&
\examplewithclass{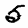}{5}&
\examplewithclass{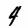}{4}&
\examplewithclass{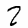}{2}&
\fbox{\examplewithclass{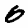}{0}}&
\examplewithclass{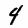}{4}& 
\\
\end{tabular}
}

\caption[Difficult MNIST test samples]{Difficult examples from
  the MNIST test set along with their target labels. At least one of the four
  state-of-the-art systems (cp.~Table~\ref{tab:mnist-er}) misclassifies these
  images.  The framed examples are misclassified by all four systems.
\label{fig:mnist-difficult}}
\end{figure}

Figure~\ref{fig:mnist-difficult} shows the `difficult' examples from the MNIST
test set. At least one of the four state-of-the-art systems misclassifies each
sample.  (These systems are marked with `$^*$' in Table~\ref{tab:mnist-er}.)
Those samples that are misclassified by all four systems are marked by a
surrounding frame. This presentation is possible because both in \cite{sch02}
and in \cite{shapecontext_pami} the authors present the set of samples
misclassified by their systems.  Furthermore, Patrice Simard kindly provided
the classification results of his system as described in \cite{simardICDAR03}
for all test data.  The availability of these results also makes it possible
to determine the error rate of a hypothetical system that combines these four
best systems as described in the following Section~\ref{sec:mnist-sota}.

Some of the images in Figure~\ref{fig:mnist-difficult} are a good illustration
of the inherent class overlap that exists for this problem: some instances of
\eg `3'~vs.`5', `4'~vs.`9', and `8'~vs.~`9' are not distinguishable by taking
into account the observed image only.  This suggests that we are dealing with
a problem with non-zero Bayes error rate. Further improvements in the error
rate on this data set might therefore be problematic. For example, consider a
classifier that classifies the second framed image as a `9': despite the fact
that this classifier would not make an error with this decision according to
the class labels, we might prefer a classifier that classifies the image as a
`4'. Note that recently~\cite{suen-prl05} has presented a more detailed
discussion of different types of errors made by state-of-the-art classifiers
for handwritten characters.

\section{The classifiers and their combination}
\label{sec:mnist-sota}

We briefly describe the four systems for handwritten digit recognition that we
compare and combine.  Then, we discuss the statistical significance of their
results and present a simple classifier combination of these four methods that
achieves a (hypothetical) error rate of 0.35\%.

{\bf Shape context matching.}  \cite{shapecontext_pami} presents the shape
context matching approach.  The method proceeds by first extracting contour
points of the images. In the case of handwritten character images the
resulting contour points trace both sides of the pen strokes the character is
composed of.  Then, at each contour point a local descriptor of the shape as
represented by the contour points is extracted. This local descriptor is
called a shape context and is a histogram of the contour points in the
surrounding of the central point. This histogram has a finer resolution at
points close to the central point and a coarser for regions farther away,
which is achieved using a log-polar
representation.\\
The classification is then done by using a nearest neighbor classifier
(although the authors chose to use only one third of the training data for the
MNIST task). The distance within the classifier is determined using an
iterative matching based on the shape context descriptors and two-dimensional
deformation. The shape contexts of training and test image are assigned to
each other by using the Hungarian algorithm on a bipartite graph
representation with edge weights according to the similarity of the shape
context descriptors. This assignment is then used to estimate a
two-dimensional spline transformation best matching the two images.  The
images are transformed accordingly and the whole process (including extraction
of shape contexts) is iterated until a stopping criterion is reached. The
resulting distance is used in the
classifier. \\
Recently, \cite{athi05} discuss a cascading technique to speed up the slow
nearest neighbor matching by ``two to three orders of magnitude''. While the
result that this discussion is based on only used the first 20,000 training
samples for reasons of efficiency and resulted in an error rate of 0.63\%
\cite{shape_context}, \cite{athi05} report an error rate of 0.54\% for the
full training set and 0.58\% for the cascaded classifier that uses only about
300 distance calculations per test.

{\bf Invariant support vector machine.} 
\cite{sch02} presents a support vector machine (SVM) that is especially suited
for handwritten digit recognition by incorporating prior knowledge about the
task. This is achieved by using virtual data or a special kernel function
within the SVM. The special kernel function applies several transformations to
the compared images that leave the class identity unchanged and return the
kernel function of the appropriate pair of transformed images. This method is
referred to as kernel jittering. The second uses so-called virtual support
vectors. This approach consists of first training a support vector machine.
Now,  the set of support vectors contains
sufficient information about the recognition problem and can therefore be
considered a condensed representation of the training data for discrimination
purposes. The method proceeds to create transformed versions of the support
vectors, which are the virtual support vectors. In the experiments leading to
the error rate of 0.56\% the transformations used were image shifts within the
eight-neighborhood plus horizontal and vertical shifts of two pixels, thus
resulting in $9+4=13$ virtual support vectors for each original support
vector. (This experiment also used the deslanted version of the MNIST data
\cite{lecun98}.) On this new set of virtual support vectors, another support
vector machine was trained and evaluated on the test set.

{\bf Pixel-to-pixel image matching with local contexts.}
\cite{icpr04_nlmatch} presents deformable models for handwritten character
recognition. It is shown that a simple zero-order matching approach called
image distortion model (IDM) can lead to very competitive results if the local
context of each pixel is considered in the distortion. The local context is
represented by a $3\times3$ surrounding window of the horizontal and vertical
image gradient, resulting in an 18-dimensional descriptor.  The IDM allows to
choose for each pixel of the test image the best fitting counterpart of the
reference image within a suitable corresponding range.  The distance as
determined by the best match between two images is then used within a
3-nearest-neighbor classifier.  More elaborate models for image matching are
also discussed, but only small improvements can be obtained at the cost of
much higher computational costs.  The IDM can be seen as the best compromise
between high classification speed and high recognition accuracy while being
conceptually very simple and easy to implement.

{\bf Convolutional neural net and virtual data.}
\label{sec:idm}
\cite{simardICDAR03} presents a large convolutional neural network of about
3,000 nodes in five layers that is especially designed for handwritten
character classification. The new concept in the approach is to present a new
set of virtual training images to the learning algorithm of the neural net in
each iteration of the training. The virtual training set is constructed from
the given training data by applying a separate two-dimensional random
displacement field that is smoothed with a Gaussian filter to each of the
images. This makes it possible to generate a very large amount of virtual data
in the order of 1,000 virtual samples for each original element of the
training data set. The data is generated on the fly in each training iteration
and therefore does not have to be saved, which avoids the problems with data
handling. Apart from the generation of virtual examples there is another point
where prior knowledge about the task comes into play, namely the use of a {\em
  convolutional} neural net. This architecture, which is described in greater
detail in \cite{lecun98}, contains prior knowledge in that it uses tying of
weights within the neural net to extract low-level features from the input
that are invariant with respect to the position within the image, and only in
later layers of the neural net the position information is used.

{\bf Discussion and combination.}
We can observe that all four methods take special measures to deal
with the image variability present in the images, using virtual data
and image matching methods. At the same time the concrete
classification algorithm seems to play a somewhat smaller role in the
performance as nearest neighbor classifiers, support vector machines,
and neural networks all perform very well. Only a slight advantage of
the neural net can be seen in the possibility to use very large
amounts of virtual data in training because the training proceeds in
several iterations, which need not use the same data but can use
distorted samples of the images instead.

Figure~\ref{fig:mnist-difficult} shows all the errors made by one of the four
classifiers. It is remarkable that only eight samples are classified
incorrectly by all four systems. This observation naturally suggests the use
of classifier combination to further reduce the error rate.  The availability
of the results of the other classifiers makes it possible to determine this
error rate of a simple hypothetical combined system.  

However, we are somewhat restricted for the choice of combination scheme,
because for two of classifiers we only know if the result was correct or not.
We thus decided to use a simple majority vote combination based on the four
classifiers, where the neural net classifier is used for tie-breaking (because
it has the best single error rate).  Note that the result is only an upper
bound of the error rate that a real combined system would have, because we do
not use the class labels the patterns were assigned to (but only the
information if the decision was correct or not). This means that in case of a
disagreement between the falsely assigned classes we could have a correct
assignment when using the class labels. Furthermore, it seems likely that the
use of the confidence values of the component classifiers in the combination
scheme could also improve the joint decision.

Using the described hypothetical combination, the resulting error rate is
0.35\%\label{mnistres}.  In the following section we will show that this
improvement has a probability of 94\% to be an improvement that is not based
on chance alone but constitutes a real improvement.

\section{Statistical analysis of results}

\begin{table}[tb]
  \caption[Significance of improvements for best MNIST classifiers]%
  {Probabilities of improvement for all pairs of the four used
  classifiers and their combination according to a bootstrap analysis. 
  Probabilities in
  {boldface} show {\bfseries significant} improvements with respect to the
  5\% level. This table can be read as follows: the classifier in
  each row improves over the classifiers given in the columns with the
  stated probability (\eg the probability of improvement for SVM over SC is
  0.60). The second table shows the difference in error rates for
  comparison.  SC: shape context matching; SVM: invariant support 
  vector machine; IDM: image distortion model; CNN: convolutional neural net with
  distortions; CC: combination of the four classifiers;}
  \label{tab:poi-mnist}
  \small
  \centering
  \begin{tabular}{|l|c|c|c|c|c|}
    \multicolumn{6}{c}{probability of improvement}\\
    \hline
        & SC & SVM & IDM & CNN & CC \\
    \hline
    SC  & --- &   &   &  &\\
    \hline
    SVM & 0.60 & --- &   & & \\
    \hline
    IDM & 0.85 & 0.58 & --- & & \\
    \hline
    CNN & {\bf 0.99} & {\bf 0.96} & 0.92 & --- &\\
    \hline
    CC  & {\bf 1.00} & {\bf 1.00} & {\bf 1.00} & 0.94  & ---  \\
    \hline
  \end{tabular} \hspace{1cm}
  \begin{tabular}{|l|c|c|c|c|c|}
    \multicolumn{6}{c}{difference in error rate}\\
    \hline
        & SC & SVM & IDM & CNN & CC\\
    \hline
    SC  & ---  &     &   &  & \\
    \hline
    SVM & 0.07 & --- &   & &  \\
    \hline
    IDM & 0.09 & 0.02 & --- & &  \\
    \hline
    CNN & 0.21 & 0.14 & 0.12 & --- & \\
    \hline
    CC  & 0.28 & 0.21 & 0.19 & 0.07 & ---  \\
    \hline
  \end{tabular}
\end{table}

As mentioned above, we can perform a more detailed analysis of the results of
the four methods described in the previous section because we do not only
know the error rate of the classifiers but also the exact patterns for which
an error has occurred. Therefore, we do not have to assume that the
classifiers have been evaluated on independent data and are thus able to 
derive tighter estimates of the level of confidence of an improvement.

The more detailed analysis shown here is an estimation of the
probability that a classifier performs generally better than a second
classifier (probability of improvement) by using the decisions of the
two classifiers on the same test samples. We estimate this probability
by drawing a large number of bootstrap samples from the test data set
and observing the relative performance of the two classifiers on these
resampled test sets~\cite{bisani_poi}. This estimation tells us more
than just using a comparison based on the individual error rates
alone. For example, we will intuitively be more inclined to believe
that the first classifier is better if it leads to better
classifications on 2\% of the test data and to the same results on the
remaining 98\% than if the first classifier performs better on 30\% of
the test data but worse on 28\% of the data. (For an interesting
discussion of significance in the context of comparisons of machine
learning algorithms, see \cite{salzberg97comparing}.)
Table~\ref{tab:poi-mnist} shows the probabilities of improvement based
on this technique for the four methods described above along with the
differences in error rate. \cite{lecun98} states that improvements of
more than 0.1\% in the error rate may be considered significant. The
analysis performed here allows a more detailed assessment of the
significance of improvements.

We observe that the improvements between the three classifiers based on shape
context, virtual support vectors, and the image distortion model, do not
differ statistically significantly (at the 5\% level). On the other hand, the
neural net based classifier shows significant improvements over the
classifiers based on shape context and virtual support vectors, but not over
the classifier based on the image distortion model. Finally, the improvements
of the combined classifier over the single classifiers is highly significant
except for the improvement with respect to the neural net, where the
improvement has a significance level of 6\%. This value is not beneath the 
commonly used 5\% threshold, but sufficiently close to it to convince 
us that the improvement is not based on chance alone.

\section{Conclusion}

We presented a statistical analysis of the results of four state-of-the-art
systems for handwritten character recognition on the MNIST benchmark.  By
using the fact that the systems were tested on the same data, we were able to
derive more specific results than it would have been possible by using the
error rates (and number of tests) alone. During the analysis, we observed that
the four systems had a higher variability in the results than we initially
expected. Specifically, only eight errors were common among all classifiers.
This observation motivated a combination of the classifiers, which resulted in
an error rate of 0.35\%, the lowest error rate reported on this data set so
far. The statistical analysis resulted in a probability of improvement of 94\% 
for the combination with respect to the best single classifier.

In the view of the low error rates that are achieved by current methods on the
MNIST data, we may have reached a point at which further improvement may be
largely due to random effects and overadaptation to the (test) data. Some of
the errors observed also show that the Bayes error rate of the problem is also
larger than zero. This underlines the necessity to present statistical
analyses of improvement claims and the measures taken to avoid training on the
testing data within all publications using these data in the future.  These
results may also be viewed as a hint that it is necessary to promote benchmark
data sets of similar impact as the MNIST data for new and more complex
problems.

\section*{Acknowledgements}
This work was partially funded by the BMBF (German Federal Ministry of
Education and Research), project IPeT (01~IW~D03).

\section*{Appendix}
For completeness, we list the numbers of the MNIST test patterns that 
are misclassified by the four systems and their combination in this appendix.

\begin{description}
\small
\addtolength{\itemsep}{-1ex}
\item[Shape Context \cite{shapecontext_pami}]
210, 448, 583, 692, 717, 948, 1034, 1113, 1227, 1248, 1300, 1320, 1531, 1682, 1710, 1791, 1879, 1902, 2041, 2074, 2099, 2131, 2183, 2238, 2448, 2463, 2583, 2598, 2655, 2772, 2940, 3063, 3074, 3251, 3423, 3476, 3559, 3822, 3851, 4094, 4164, 4202, 4370, 4498, 4506, 4663, 4732, 4762, 5736, 5938, 6555, 6572, 6577, 6598, 6884, 8066, 8280, 8317, 8528, 9506, 9643, 9730, 9851
\item[SVM \cite{sch02}]
448, 583, 660, 675, 727, 948, 1015, 1113, 1227, 1233, 1248, 1300, 1320, 1531, 1550, 1682, 1710, 1791, 1902, 2036, 2071, 2099, 2131, 2136, 2183, 2294, 2489, 2655, 2928, 2940, 2954, 3031, 3074, 3226, 3423, 3521, 3535, 3559, 3605, 3763, 3870, 3986, 4079, 4762, 4824, 5938, 6577, 6598, 6784, 8326, 8409, 9665, 9730, 9750, 9793, 9851
\item[IDM \cite{icpr04_nlmatch}]
446, 448, 552, 717, 727, 948, 1015, 1113, 1243, 1682, 1879, 1902, 2110, 2131, 2183, 2344, 2463, 2524, 2598, 2649, 2940, 3226, 3423, 3442, 3559, 3602, 3768, 3809, 3986, 4054, 4164, 4177, 4202, 4285, 4290, 4762, 5655, 5736, 5938, 6167, 6884, 7217, 8317, 8377, 8409, 8528, 9010, 9506, 9531, 9643, 9680, 9730, 9793, 9851
\item[Neural Net \cite{simardICDAR03}]
583, 948, 1233, 1300, 1394, 1879, 1902, 2036, 2131, 2136, 2183, 2463, 2583, 2598, 2655, 2928, 2971, 3289, 3423, 3763, 4202, 4741, 4839, 4861, 5655, 5938, 5956, 5974, 6572, 6577, 6598, 6626, 8409, 8528, 9680, 9693, 9699, 9730, 9793, 9840, 9851, 9923
\item[Combination] 
448, 583, 948, 1113, 1233, 1300, 1682, 1879, 1902, 2036, 2131, 2136, 2183, 2463, 2583, 2598, 2655, 2928, 2940, 3423, 3559, 3763, 4202, 4762, 5655, 5938, 6572, 6577, 6598, 8409, 8528, 9680, 9730, 9793, 9851
\end{description}

\end{document}